\documentclass[letterpaper]{article} % DO NOT CHANGE THIS
\usepackage{aaai2026}  % DO NOT CHANGE THIS
\usepackage{times}  % DO NOT CHANGE THIS
\usepackage{helvet}  % DO NOT CHANGE THIS
\usepackage{courier}  % DO NOT CHANGE THIS
\usepackage[hyphens]{url}  % DO NOT CHANGE THIS
\usepackage{graphicx} % DO NOT CHANGE THIS
\usepackage{natbib}  % DO NOT CHANGE THIS AND DO NOT ADD ANY OPTIONS TO IT
\usepackage{caption} % DO NOT CHANGE THIS AND DO NOT ADD ANY OPTIONS TO IT
\usepackage{algorithm}
\usepackage{algorithmic}
\usepackage{amsmath, amssymb}
\usepackage{booktabs}
\usepackage{multirow}
\usepackage{newfloat}
\usepackage{listings}
\DeclareCaptionStyle{ruled}{labelfont=normalfont,labelsep=colon,strut=off} % DO NOT CHANGE THIS
\floatstyle{ruled}
\newfloat{listing}{tb}{lst}{}
\floatname{listing}{Listing}
\nocopyright 
\title{Eigen Neural Network: Unlocking Generalizable Vision with Eigenbasis}
\author{
    Anzhe Cheng\textsuperscript{\rm 1}
    Chenzhong Yin\textsuperscript{\rm 1}
    Mingxi Cheng\textsuperscript{\rm 1}
    Shukai Duan\textsuperscript{\rm 1}\\
    Shahin Nazarian\textsuperscript{\rm 1}
    Paul Bogdan\textsuperscript{\rm 1}\\
}
\affiliations{
    \textsuperscript{\rm 1} University of Southern California
}

\usepackage{bibentry}
\begin{document}

\maketitle

\begin{abstract}
The remarkable success of Deep Neural Networks(DNN) is driven by gradient-based optimization, yet this process is often undermined by its tendency to produce disordered weight structures, which harms feature clarity and degrades learning dynamics. To address this fundamental representational flaw, we introduced the Eigen Neural Network (ENN), a novel architecture that reparameterizes each layer's weights in a layer-shared, learned orthonormal eigenbasis. This design enforces decorrelated, well-aligned weight dynamics axiomatically, rather than through regularization, leading to more structured and discriminative feature representations. When integrated with standard BP, ENN consistently outperforms state-of-the-art methods on large-scale image classification benchmarks, including ImageNet, and its superior representations generalize to set a new benchmark in cross-modal image-text retrieval. Furthermore, ENN's principled structure enables a highly efficient, backpropagation-free(BP-free) local learning variant, ENN-$\ell$. This variant not only resolves BP's procedural bottlenecks to achieve over 2$\times$ training speedup via parallelism, but also, remarkably, surpasses the accuracy of end-to-end backpropagation. ENN thus presents a new architectural paradigm that directly remedies the representational deficiencies of BP, leading to enhanced performance and enabling a more efficient, parallelizable training regime.
\end{abstract}


\section{Introduction}
\label{sec:intro}

Backpropagation(BP), while foundational to deep learning, suffers from fundamental limitations that compromise both training efficiency and biological plausibility. The BP training process frequently leads to misaligned weight structures that harm feature clarity and learning dynamics~\cite{szegedy2013intriguing,lillicrap2020backpropagation}. 
A central limitation of backpropagation is that its weight updates are largely unstructured, yielding highly redundant parameterizations: in standard MLPs and convnets, over 95\% of weights can be accurately predicted from a small subset without harming accuracy~\cite{denil2013predicting}. 
More recent studies confirm this dynamic. Forward and backward weights trained with BP often remain disordered, leading to ill-conditioned dynamics and slower convergence~\cite{flugel2024feed,ji2024deep}, and even architectures with random feedback require soft alignment at initialization for stable convergence~\cite{cheon2025one}. For example, Frozen Backpropagation~\cite{goupy2025frozen} reports that a ResNet-50 trained on ImageNet retains a forward–feedback cosine alignment below 0.2 after 50 epochs, stalling optimization despite periodic re-synchronization of feedback weights. These findings indicate that standard BP produces feature maps dominated by redundant or entangled directions, especially in deep networks.

Beyond weight disorder, BP also enforces strict layer-wise dependencies. Each layer’s weight update must wait until gradients are back-propagated from the output and through all subsequent layers, a phenomenon known as backward-locking; likewise, no weight can update until the full forward/backward pass is completed (update-locking)~\cite{huo2018training,jaderberg2017decoupled}.
These locking constraints, together with the neurobiologically implausible requirement of symmetric forward and feedback weights (referred to as the weight transport problem~\cite{liao2016important,shervani2023meta}), introduce significant inefficiencies. Specifically, they slow down training, increase memory consumption, and hinder on-the-fly learning.
For instance, in a CLIP-based interactive image-retrieval system~\cite{wang2025fix,zhang2024long}, incorporating real-time user feedback would require a full backpropagation pass through hundreds of millions of parameters, an impractical task given BP’s need for global synchronization.

\begin{figure*}[t]
\centering
\includegraphics[width=\linewidth]{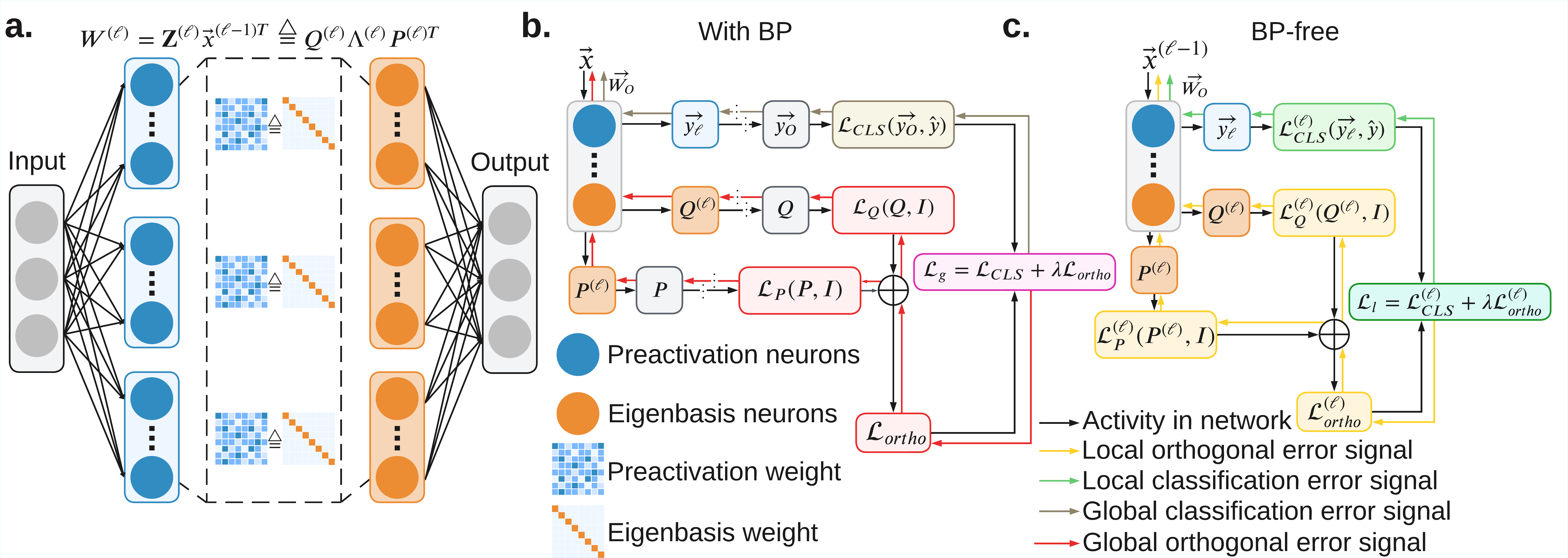}
\caption{\textbf{Overview of the Eigen Neural Network (ENN) framework.} (a) Eigenvector parameterization. Layer $\ell$ decomposes its preactivation weight matrix into an eigenbasis. $Z^{(\ell)}$ collects pre-activation coefficients, $\vec{x}^{(\ell-1)}$ is the input from last layer. $Q^{(\ell)}$ and $P^{(\ell)}$ are orthonormal basis, $\Lambda^{(\ell)}$ is a learned diagonal vector. (b) In a conventional BP pipeline, global classification loss $\mathcal{L}_{\text{CLS}}$ and orthonormality loss $\mathcal{L}_{\text{ortho}}$ are minimized via chained gradients. The ellipsis (\dots) on each arrow signals that the activation continues through all subsequent layers to the final output. $\hat{y}$ is the ground truth, $\vec{y}_{O}$ is the global output. (c) In ENN’s BP-free mode, each layer predicts locally with logits $\vec{y}^{(\ell)}$ and receives a local error signal $\mathcal{L}_\ell$, enabling weight updates that are fully parallel and require no weight transport or global synchronization.}
\label{fig:enn-architecture}
\end{figure*} 

To address these fundamental limitations of BP, we introduced Eigen Neural Networks (ENN) – a novel neural architecture and training paradigm designed for more effective and parallelizable learning. ENN reformulates network weights in a shared eigenbasis: instead of learning each weight parameter directly, the model learns a set of 
orthonormal eigenbasis vectors and locally trainable coefficients that linearly combine those basis vectors to form the weights of each layer. 
In other words, each layer’s weight matrix is expressed as a weighted sum of common eigenbasis components. This reparameterization directly targets BP’s inefficiencies. First, by learning in the eigenbasis, ENN ensures that all trainable coefficients receive meaningful gradient signals – no component of the weight space is left with vanishing or neglected updates. The orthonormal eigenbasis provides well-conditioned directions for error propagation, so gradient information is distributed across all coefficients without dying out, eliminating gradient starvation. Second, ENN’s weight update mechanism breaks the backward/update locking of standard BP. Since each layer’s weight coefficients are updated in a decoupled eigenbasis, they can be optimized simultaneously and independently. Layers no longer need to wait for sequential backpropagation; each can apply gradient updates in parallel, free of the global locking that slows ordinary BP. 
Hence, the ENN architecture naturally integrates into BP-free, block-wise frameworks—referred to as ENN-$\ell$—significantly enhancing their parallelism and performance.

The primary contributions of this research can be summarized as follows:
\begin{itemize}
    \item We introduced Eigen Neural Networks (ENN), a general-purpose deep learning architecture that represents synaptic weights in a shared eigenvector basis, fundamentally altering how weights are parameterized and coupled within each layer.
    \item We embedded the ENN model on top of recent BP-based neural network architectures, significantly improving their performance on both image classification (CIFAR-10, CIFAR-100, Tiny ImageNet, and ImageNet) and cross-modal image-text retrieval tasks (COCO and Flickr30K). 
    % ~\textcolor{blue}{add some parameters here}
    \item Remarkably, our ENN-$\ell$ surpasses leading BP-free baselines and achieves comparable accuracy to BP training when run on the same backbones. Moreover, ENN-$\ell$ attains over $2\times$ speedup compared to BP with the same backbone, demonstrating superior parallelism and computational efficiency. 
\end{itemize}

\section{Related Work}

Deep vision models such as ResNet~\cite{he2016deep} and Vision Transformer (ViT)~\cite{dosovitskiy2020image} owe their success to error backpropagation, but this comes with low training efficiency and biological implausibility. Consequently, two main lines of research have emerged to address BP's limitations: those that improve its representational quality through regularization, and those that replace it entirely with BP-free learning rules. 

Regularization methods aim to impose a beneficial structure within the BP framework. For instance, DropCov~\cite{wang2022dropcov} uses adaptive dropout to decorrelate features, while orthogonality-enforcing methods like OCNN~\cite{wang2020orthogonal} and SRIP~\cite{bansal1810can} regularize weight matrices to improve feature diversity. Pushing this further, DeCEF~\cite{yu2024building} factorizes convolutions into shared eigen-filters to impose a low-rank structure. While these techniques improve feature quality without increasing parameter counts, they remain fundamentally constrained by BP's procedural bottlenecks, leaving the global, sequential gradient flow untouched.

To overcome these procedural flaws, BP-free strategies avoid global gradient propagation altogether. One approach is to train networks in a block-wise fashion. Block-Local Learning (BLL)~\cite{fokamvariational} and BWBPF~\cite{cheng2024unlocking} partition the network into modules and apply standard backpropagation within each block using locally generated targets, thereby reducing inter-layer dependencies. Similarly, COMQ~\cite{zhang2025comq} formulates post-training quantization as a purely local, BP-free calibration process, allowing each layer to be quantized using its own error metric. While these block-based methods alleviate gradient locking and reduce memory usage, they often require careful block design and can show diminishing returns as model depth increases. Overall, existing BP-Free methods improve efficiency and biological plausibility but still face a trade-off in terms of accuracy, stability, or scalability, motivating the need for a unified solution.

\section{Methodology}
\label{sec:method}

Our ENN model replaces conventional weight matrices with an eigenvector-based parameterization to enforce a structured and decorrelated feature space. As illustrated in Figure~\ref{fig:enn-architecture}a, each layer $\ell$ in the network maps its incoming activations $\vec{x}^{(\ell-1)}$ to an output preactivation vector $\mathbf{Z}^{(\ell)}$. In a standard feedforward layer, this mapping is linear, which is denoted as:
\begin{equation}
    \mathbf{Z}^{(\ell)} = W^{(\ell)}\vec{x}^{(\ell-1)},
\end{equation}

where $W^{(\ell)}$ is the weight matrix for layer $\ell$. Equivalently, for a given input-output pair, the weight could be expressed as:
\begin{equation}
    W^{(\ell)} = \mathbf{Z}^{(\ell)}\vec{x}^{(\ell-1)T},
\end{equation}

In other words, $W^{(\ell)}$ linearly transforms the input $\vec{x}^{(\ell-1)}$ into the layer’s preactivation space, which consists of the preactivation neurons. This forms the basis of the conventional architecture shown in Figure~\ref{fig:enn-architecture}a, mapping the network input through successive linear weights and nonlinearities to produce the final output.
 
\textbf{Eigenvector-based Weight Parameterization.} At the core of our approach is an eigenbasis weight reparameterization that factorizes each weight matrix into orthogonal bases and a diagonal scaling. Specifically, we represent each layer’s weight as:
\begin{equation}
W^{(\ell)} = Q^{(\ell)} \Lambda^{(\ell)} P^{{(\ell)}T},
\end{equation}

where $Q^{(\ell)}$ and $P^{(\ell)}$ are orthonormal matrices, and $\Lambda^{(\ell)}$ is a diagonal matrix of the same shape as $W^{(\ell)}$ with nonzero entries only on its main diagonal. The columns of $P^{(\ell)}$ and $Q^{(\ell)}$ can be thought of as orthogonal eigenbasis neurons for the layer’s input and output spaces, respectively. Intuitively, $P^{(l)T}$ projects the input $\vec{x}^{(l-1)}$ into this shared eigenbasis, $\Lambda^{(l)}$ scales the resulting coordinates, and $Q^{(l)}$ maps the scaled vector into the output preactivation space. This decomposition is applied to all layer types by reshaping their weight tensors into matrices. Importantly, this structured representation does not add trainable parameters, as the factors $Q^{(l)}$, $\Lambda^{(l)}$, and $P^{(l)}$ have the same degrees of freedom as the original $W^{(l)}$, with $Q^{(l)}$ and $P^{(l)}$ constrained to remain orthonormal.

\textbf{ENN with Global Backpropagation.} To overcome the drawbacks of disorder weights in traditional training, we integrate our reparameterized layers into a standard BP training framework as shown in Figure~\ref{fig:enn-architecture}b. In this BP-integrated version of ENN, the network is trained end-to-end with a global objective function, $\mathcal{L}_{g}$, which combines a primary task loss with an orthogonal loss term to maintain the eigenbasis structure. Take classification as an example, the objective is:
\begin{equation}
    \mathcal{L}_{g} = \mathcal{L}_{CLS} + \lambda\mathcal{L}_{ortho},
    \label{eq:lg}
\end{equation}
where $\mathcal{L}_{CLS}$ is the standard classification loss (e.g., cross-entropy) computed on the final network output, which updates the whole weight(i.e., $Q$, $\Lambda$, $P$) throughout the training process, and $\mathcal{L}_{ortho}$ is a global orthogonality penalty. This penalty encourages the basis matrices in every layer to remain orthonormal throughout training and is defined as the sum of the squared Frobenius norms~\cite{bottcher2008frobenius} of their deviation from orthogonality:

\begin{equation}
L_{\mathrm{orth}}
\;=\;
\bigl\lVert Q^{\top} Q - I \bigr\rVert_{F}^{2}
\;+\;
\bigl\lVert P^{\top} P - I \bigr\rVert_{F}^{2},
\end{equation}
which is zero if $Q$ and $P$ are perfectly orthonormal to themselves. This term penalizes any drift in the basis vectors away from orthogonality as the network learns, thereby maintaining the validity of the $W^{(\ell)} = Q^{(\ell)} \Lambda^{(\ell)} P^{(\ell)T}$ factorization. 

The hyperparameter $\lambda$ in Equation(\ref{eq:lg}) balances the two loss components. The $\lambda$ is chosen to be small so that preserving orthonormality does not unduly interfere with the layer’s primary task of minimizing classification error. In our experiment setting, we set $\lambda=2\times10^{-4}$. (Details on selecting $\lambda$ in our experiment are provided in the Appendix.) Both error signals are global, and their gradients are backpropagated through all layers to update the eigenbasis parameters ($Q^{(l)}, \Lambda^{(l)}, P^{(l)}$) of each layer.

\textbf{ENN-$\ell$ with Local Learning.} We also proposed a BP-free training regime for ENN, denoted as ENN-$\ell$, which is outlined in Figure~\ref{fig:enn-architecture}c. In this BP-free version, each layer is trained independently using purely local objectives, without receiving any error gradients from other layers. To achieve this, we attach a lightweight classifier to each layer $\ell$ that produces a local prediction $\vec{y}_\ell$ for the final task. This local head takes the layer’s activations and attempts to predict the true label. We define a layer-specific loss for each layer $\ell$ as:

\begin{equation}
    \mathcal{L}_{l} = \mathcal{L}^{(l)}_{CLS} + \lambda \mathcal{L}^{(l)}_{ortho}, 
\end{equation}

Here, $\mathcal{L}^{(l)}_{CLS}$ is the classification loss computed at layer $\ell$’s local output $\vec{y}_\ell$ against the global ground-truth label $\hat{y}$, and $\mathcal{L}^{(l)}_{ortho}$ is a layer-specific orthonormality penalty for the eigenbasis weights of that layer.This loss is analogous to the global $\mathcal{L}_{ortho}$, but applied to $Q^{(\ell)}$ and $P^{(\ell)}$. Each layer minimizes its own loss $\mathcal{L}_{\ell}$ independently. In the forward pass, the network’s layers still feed into one another, so that deeper layers receive increasingly processed representations. But during learning, no gradient signals flow backward from layer $\ell+1$ to $\ell$. In other words, layer $\ell$ is trained to improve its own local prediction $\vec{y}_\ell$ and maintain orthonormal weights without any direct knowledge of the final output or the losses of other layers. Each layer receives the input, produces an output and a prediction, and updates its parameters based only on its local error $\mathcal{L}_{\ell}$. 

\begin{table}[t]
\centering
\setlength{\tabcolsep}{2.7pt} 
\small
\begin{tabular}{c|ccc|ccc}
\toprule
\multirow{3}{*}{\textbf{Method}} & 
\multicolumn{3}{c|}{\textbf{CIFAR-10}} &
\multicolumn{3}{c}{\textbf{CIFAR-100}} \\ 
& ResNet & ResNet & ResNet
& ResNet & ResNet & ResNet \\ 
&50&101&152
&50&101&152\\\hline

BP                & 7.99 & 7.14 & 6.35 & 32.94 & 30.70 & 29.18 \\
OCNN    & 7.79 & 7.57 & 6.40 & 26.47 & 25.69 & 22.06\\
DeCEF   & 7.13 & 7.07 & 6.63 & 28.24 & 25.16 & 22.20\\
DropCov &  \underline{7.11} &\underline{7.09} &6.65 &28.74 &\underline{24.96} &\underline{20.94}\\
SRIP   & 7.87 & 7.55 & \underline{6.28} & \underline{26.03} & 25.14 & 21.94 \\\hline
\textbf{ENN}         & \textbf{7.06} & \textbf{7.02} & \textbf{5.96} & \textbf{25.78} & \textbf{24.82} & \textbf{20.20} \\ 
\bottomrule
\end{tabular}
\caption{Test classification error (\% $\downarrow$) for ENN and advanced BP image classification models with ResNet backbones on CIFAR datasets.
\textbf{Bold} denotes the best (lowest) error and \underline{underline} the second‐best error rate.}
\label{tab:img_bp}
\end{table}

\section{Experiments}
In this section, we conducted a comprehensive and multifaceted empirical validation of the ENN framework. First, we demonstrated the profound representational benefits of ENN's eigenbasis parameterization when integrated into a standard BP pipeline, directly addressing the issue of misaligned weight structures. Second, we confirmed the superior generalization capabilities of these enhanced representations by applying them to a challenging cross-modal image-text retrieval task. Third, we showcased the transformative capabilities of the BP-free variant, ENN-$\ell$, establishing its superiority in both accuracy and computational efficiency over both BP and existing BP-free methods. Finally, we stress-test the scalability and robustness of ENN-$\ell$ on high-dimensional 3D neuroimaging data, confirming its applicability to complex scientific problems.

\subsection{Image Classification with BP-Integrated}
We first evaluated ENN's ability to improve feature representations when integrated into a conventional end-to-end BP training pipeline. This setup demonstrates the benefits of our eigenvector-based weight parameterization.

\textbf{Experimental Setup.} Our experiments span multiple datasets of increasing complexity: CIFAR-10, CIFAR-100~\cite{krizhevsky2009learning}, Tiny ImageNet~\cite{le2015tiny}, and the full ImageNet ILSVRC 2012 dataset~\cite{deng2009imagenet}. We used ResNet (50, 101, 152)~\cite{he2016deep} and Vision Transformer (ViT-B/16)~\cite{dosovitskiy2020image} backbones to demonstrate architectural generality. All models are implemented in PyTorch, trained from scratch on two NVIDIA A100 GPUs using the AdamW optimizer with standard data augmentations. Detailed hyperparameters for each dataset are provided in the Appendix. We compared ENN against standard BP and leading regularization-based methods, including OCNN~\cite{wang2020orthogonal}, DeCEF~\cite{yu2024building}, DropCov~\cite{wang2022dropcov}, and SRIP~\cite{bansal1810can}. We reported the top-1 classification error rate (\%) for all models.

\begin{figure}[t]
\centering
\includegraphics[width=\linewidth]{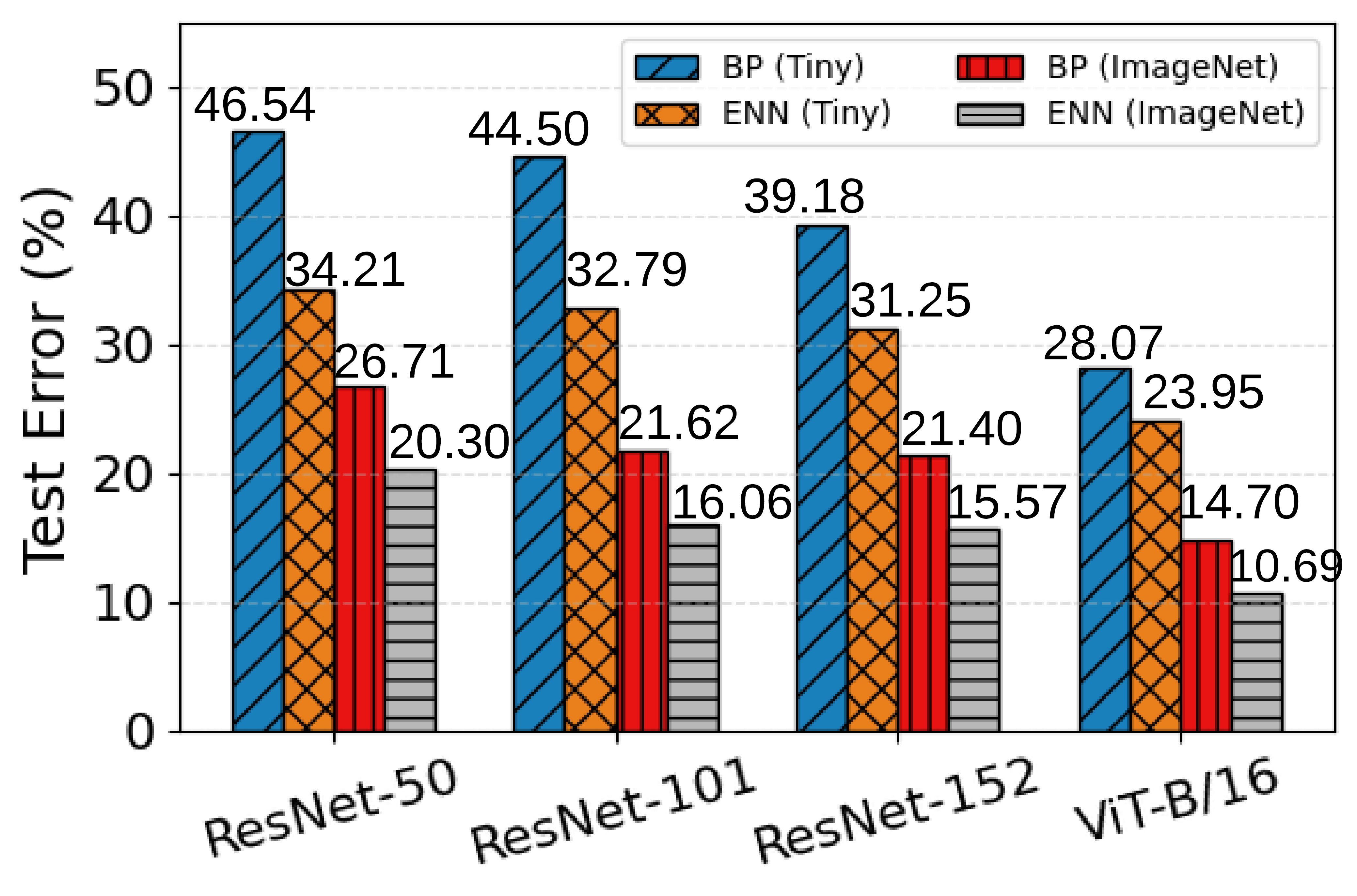}
\caption{\textbf{Test classification error rate (\%~$\downarrow$) comparison of ENN with BP on ImageNet datasets.} ENN dominate BP on such complex datasets on all backbones}
\label{fig:enn_bp_imagenet_err}
\end{figure}

\textbf{CIFAR-10 and CIFAR-100.} As shown in Table~\ref{tab:img_bp}, ENN consistently achieves the lowest test error across all ResNet architectures on both CIFAR datasets, establishing a new state-of-the-art among the tested methods. On CIFAR-10, for example, ENN with a ResNet-50 backbone reduces the error rate to 7.06\%, a notable improvement over the 7.99\% from standard BP. The performance gains are more pronounced on CIFAR-100, where ENN cuts the ResNet-50 error from 32.94\% to 25.78\%, which is a \textbf{21.7\%} relative reduction. This widening performance gap is not coincidental; it points to a core strength of the ENN architecture. Fine-grained classification tasks like CIFAR-100 demand that the model learn subtle yet critical distinctions between visually similar classes, a process that is often hindered by the feature entanglement resulting from the disordered weight updates of standard BP. By enforcing an orthonormal eigenbasis, ENN axiomatically encourages the formation of decorrelated, disentangled features. This architectural prior is thus disproportionately beneficial for learning the highly discriminative representations necessary to resolve the ambiguities inherent in fine-grained classification, leading to a more significant performance uplift as task complexity increases.

\begin{table}[t]

\centering
\begin{tabular}{lccc}
\toprule
\textbf{Method} & \textbf{ResNet-50} & \textbf{ResNet-101} & \textbf{ResNet-152} \\\hline
BP     & 26.71&21.62&21.40\\
OCNN   & 24.36 & 18.73 & 17.17 \\
DeCEF  & 23.39 & \underline{17.50} & 17.00 \\
DropCov & \underline{21.81} & 20.49 & 17.74 \\
SRIP   & 26.10 & 18.94 & \underline{16.88} \\\hline
\textbf{ENN}    & \textbf{20.30} & \textbf{16.06} & \textbf{15.57} \\
\bottomrule
\end{tabular}
\caption{Test classification error (\% $\downarrow$) for ENN and advanced BP image classification models on ImageNet.
\textbf{Bold} denotes the best (lowest) error and \underline{underline} the second‐best error rate.}
\label{tab:enn_bp_imagenet_err}
\end{table}

\begin{figure}[t]
\centering
\includegraphics[width=0.9\linewidth]{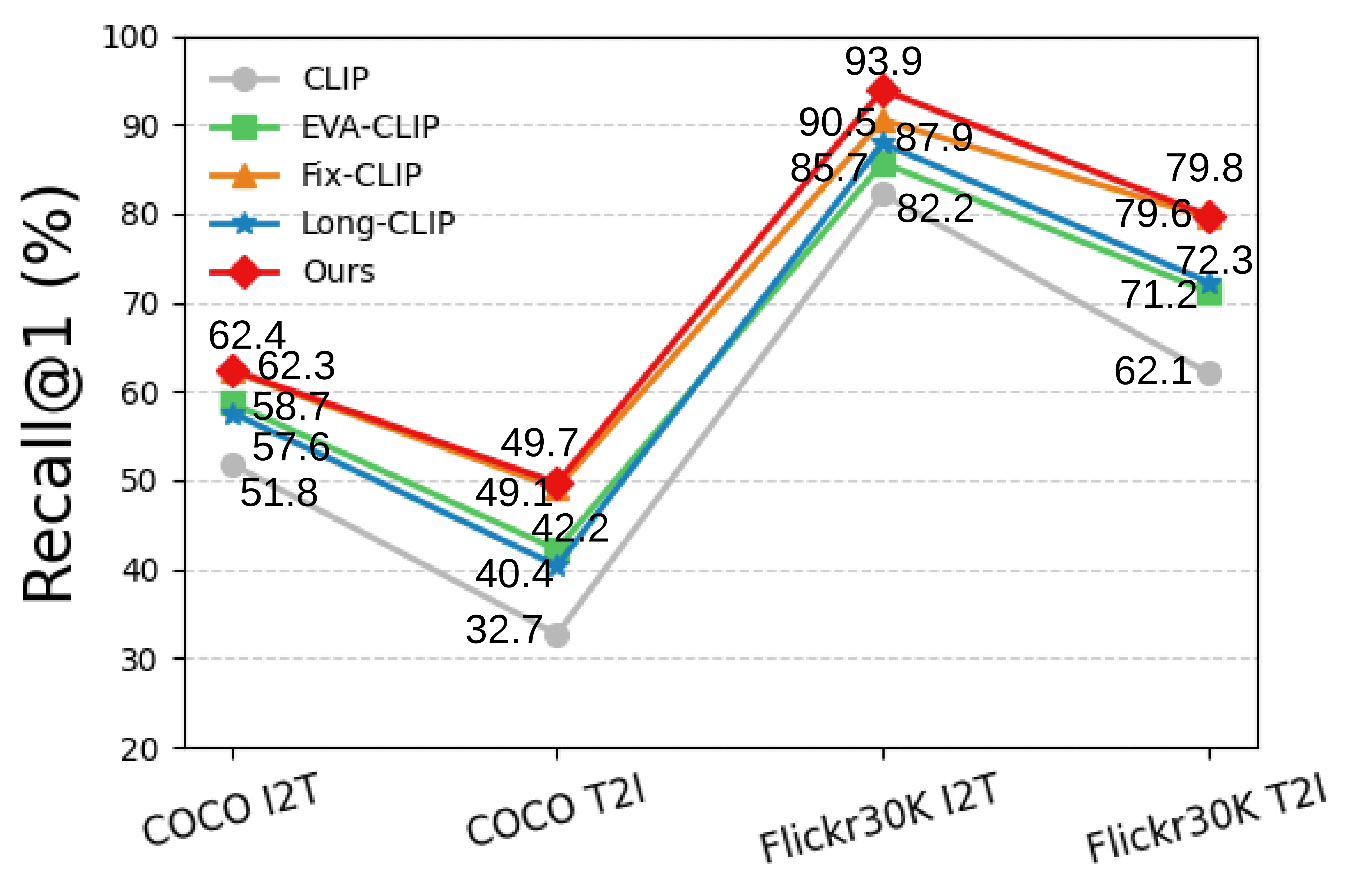}
\caption{\textbf{Comparison of our Long-ECLIP model with CLIP, EVA-CLIP, Fix-CLIP, and Long-CLIP across the four standard R@1 tasks.} Our approach shows the strongest performance on all retrieval benchmarks.}
\label{fig:radar}
\end{figure}

\begin{table*}[t]
\centering
\small
\begin{tabular}{l |ccc | ccc | ccc| ccc}
\toprule
\multirow{2}{*}{Method} &  
\multicolumn{6}{c|}{\textbf{COCO}} &
\multicolumn{6}{c}{\textbf{Flickr30K}} \\

% \cmidrule(lr){3-8}\cmidrule(lr){9-14}
& 
\multicolumn{3}{c|}{Image-to-Text} & \multicolumn{3}{c|}{Text-to-Image} &
\multicolumn{3}{c|}{Image-to-Text} & \multicolumn{3}{c}{Text-to-Image}  \\
 & R@1 & R@5 & R@10 & R@1 & R@5 & R@10 & R@1 & R@5 & R@10 & R@1 & R@5 & R@10  \\\hline
CLIP$^\dagger$          & 51.8 & 76.8 & 84.3 & 32.7 & 57.7 & 68.2 & 82.2 & 96.6 & 98.8 & 62.1 & 85.7 & 91.8 \\
EVA-CLIP$^\dagger$        & 58.7 & 80.7 & 88.2 & 42.2 & 66.9 & 76.3 & 85.7 & 96.7 & \underline{98.9} & 71.2 & 91.0 & 94.7  \\
LoLTIP$^\dagger$          & 59.7 & 81.5 &  –   & 38.1 & 63.8 &  –   & 86.9 & 97.8 &  –   & 65.2 & 88.0 &  –   \\
DreamLIP$^\dagger$       & 58.3 & 81.6 & 88.8 & 41.1 & 67.0 & 76.6 & 87.2 & 97.5 & 98.8 & 66.4 & 88.3 & 93.3 \\
Fix-CLIP$^\dagger$        & \underline{62.3} & \underline{85.4} & \underline{91.4} & \underline{49.1} & \textbf{73.8} & \textbf{82.4} & \underline{90.5} & \underline{99.0} & \textbf{99.8} & \underline{79.6} & \textbf{94.9} & \textbf{97.4}  \\
Long-CLIP$^\dagger$     &57.6 &81.1 &87.8 &40.4 &65.8 &75.2 &87.9 &97.2 &\underline{98.9} &72.3 &92.2 &95.6\\
\hline
\textbf{Long-ECLIP}     & \textbf{62.4}   & \textbf{88.1}   & \textbf{94.5}   & \textbf{49.7}   & \underline{72.6}   & \underline{82.2}   & \textbf{93.9}   & \textbf{99.3}   & \textbf{99.8}  & \textbf{79.8}   & \underline{93.5}   & \underline{97.1}  \\
\bottomrule
\end{tabular}
\caption{Bidirectional image–text retrieval Recall(\% $\uparrow$) on MS-COCO and Flickr30K. All methods use a ViT-B/16 image encoder for fair comparison. We report Recall@K for both image-to-text (I2T) and text-to-image (T2I) tasks. \textbf{Bold} denotes the best results and \underline{underline} the second‐best results. $^\dagger$Published baselines are taken from Fix-CLIP~\cite{wang2025fix}. “--” indicates the result was not reported for that metric.}
\label{tab:retrieval}
\end{table*}

\textbf{Scaling up to ImageNet and its variants.} The benefits of ENN scale effectively to large-scale, high-resolution datasets, as demonstrated in Figure~\ref{fig:enn_bp_imagenet_err} and detailed in Table~\ref{tab:enn_bp_imagenet_err}. Across all tested backbones, ENN maintains a significant performance advantage over both standard BP and the suite of advanced regularization methods. For example, with a ResNet-152 backbone, ENN achieves a top-1 error of 15.57\%, a remarkable improvement over the 21.40\% error of the standard BP-trained baseline and a clear victory over the 16.88\% from the strongest competing method, SRIP. This trend holds robustly for transformer architectures; when integrated with ViT-B/16, ENN reduces the error from 14.70\%(BP) to 10.69\%, a relative reduction of \textbf{27.3\%}. The consistent margin of improvement across both convolutional and transformer architectures, and across shallow and deep models, indicates that the eigenbasis formulation delivers a fundamental, depth-agnostic enhancement to feature extraction rather than a scale-specific artifact.

\begin{figure}[t]
\centering
\includegraphics[width=\linewidth]{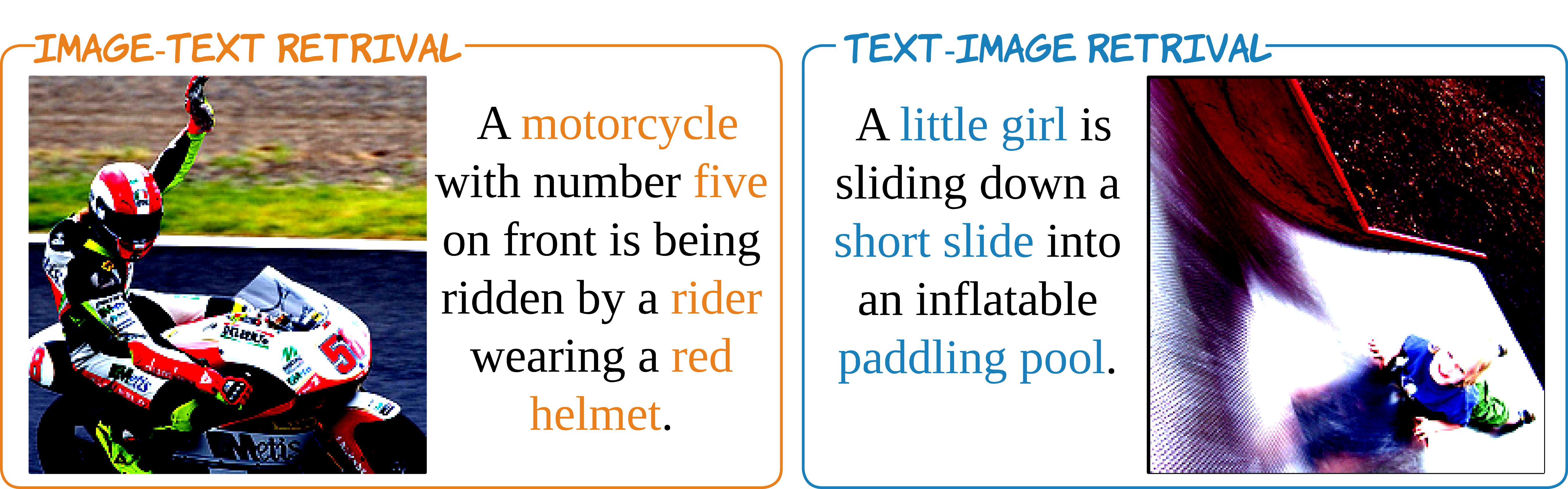}
\caption{\textbf{An example of Long-ECLIP bidirectional retrieval.} The model precisely matches a query image to its caption (orange panel) and a query caption to its image (blue panel), exemplifying ENN’s ability to encode fine-grained visual and textual attributes under the same eigenvector-trained representation.}
\label{fig:it_visual}
\end{figure}

These results, spanning diverse datasets and architectures, provide conclusive evidence that the eigenbasis reparameterization is a scalable and fundamental solution to the problem of disordered weight structures in deep networks. The significant performance gains confirm that ENN's architectural prior is not a small-scale anomaly but a robust principle that enhances representation quality at all scales.

\subsection{Cross-Modal Image-Text Retrieval}

To demonstrate that the representational benefits of ENN are not narrowly confined to classification but yield fundamentally better, general-purpose features, we evaluated ENN's transferability to the complex domain of cross-modal reasoning.

\textbf{Experimental Setup.} We adapt the Long-CLIP~\cite{zhang2024long} model by replacing its original CLIP image encoder with our ENN-trained ViT-B/16 encoder, which we denote \textbf{Long-ECLIP}. To isolate the contribution of the image representations, the text tower remains frozen with its pretrained weights. We evaluate on the MS-COCO~\cite{lin2014microsoft} and Flickr30K~\cite{plummer2015flickr30k} benchmarks, reporting standard Recall@K (R@K, K=1, 5, 10) for both image-to-text (I2T) and text-to-image (T2I) retrieval. Baselines include strong vision-language models such as CLIP~\cite{radford2021learning}, EVA-CLIP~\cite{sun2023eva}, Fix-CLIP~\cite{wang2025fix}, the original Long-CLIP and so on.

\textbf{Retrieval Performance.} As shown in Figure~\ref{fig:radar} and detailed in Table~\ref{tab:retrieval}, our ENN-enhanced model, Long-ECLIP, establishes a new state-of-the-art on both MS-COCO and Flickr30K. Specifically, Long-ECLIP outperforms all prior methods on R@1 metrics, including strong vision-language models like EVA-CLIP~\cite{sun2023eva} and Fix-CLIP~\cite{wang2025fix}, demonstrating powerful generalization.

\begin{table*}[t]
\centering
\small
\begin{tabular}{c|ccc|ccc|ccc|ccc}
\toprule
\multirow{3}{*}{\textbf{Method}} & 
\multicolumn{3}{c|}{\textbf{CIFAR-10}} &
\multicolumn{3}{c|}{\textbf{CIFAR-100}} &
\multicolumn{3}{c|}{\textbf{Tiny ImageNet}} &
\multicolumn{3}{c}{\textbf{ImageNet}} \\ 
& ResNet & ResNet & ResNet
& ResNet & ResNet & ResNet
& ResNet & ResNet & ResNet
& ResNet & ResNet & ViT \\ 
&50&101&152
&50&101&152
&50&101&152
&101&152&B/16\\\hline

BP                          & 7.99 & 7.14 & 6.35 & 32.94 & 30.70 & 29.18 & 46.54 & 44.50 & 39.18 & 21.62 & 21.40 & 14.70 \\
BLL$^{\dagger}$             & 5.40 & 5.12 & 4.98 & 32.10 & 31.74 & 30.30 & 45.06 & 44.86 & 44.33 & 20.88 & 20.69 & -- \\
SEDONA$^{\dagger}$          & 7.53 & 6.59 & 6.13 & 30.33 & 29.87 & \underline{28.67} & 45.60 & 40.88 & 35.90 & \underline{20.72} & 20.20 & -- \\
DGL$^{\dagger}$           &  8.27 &8.30 &6.39 &31.93 &31.24 &29.94 &46.04 &46.20 &42.36 &22.35 &22.20 & -– \\
BWBPF$^{\dagger}$           & \underline{4.85} & \underline{4.53} & \underline{4.48} & \underline{29.88} & \underline{29.76} & 28.79 & \underline{35.83} & \underline{35.28} & \underline{35.01} & 20.92 & \underline{20.08} & \textbf{11.16} \\
COMQ$^{\dagger}$            & 8.50 & 8.02 & 7.73 & 33.11 & 32.57 & 30.06 & 47.13 & 45.31 & 40.01 & 29.27 & 28.79 & 22.92 \\\hline
\textbf{ENN-$\ell$}         & \textbf{4.12} & \textbf{3.89} & \textbf{3.74} & \textbf{27.58} & \textbf{27.04} & \textbf{26.11} & \textbf{33.75} & \textbf{33.96} & \textbf{33.42} & \textbf{17.45} & \textbf{17.11} & \underline{11.72} \\ 
\bottomrule
\end{tabular}
\caption{Test classification error (\% $\downarrow$) for advanced image classification models with ResNet and ViT backbones.
\textbf{Bold} denotes the best (lowest) error and \underline{underline} the second‐best error rate.  
$^{\dagger}$\,Published baselines are taken from BLL~\cite{fokamvariational}, SEDONA~\cite{pyeon2020sedona}, DGL~\cite{belilovsky2020decoupled}, BWBPF~\cite{cheng2024unlocking}, and COMQ~\cite{zhang2025comq}.  
“--” indicates that the corresponding paper did not report that configuration.}
\label{tab:img_class}
\end{table*}

The most compelling evidence for ENN's contribution lies in the direct comparison with the Long-CLIP baseline. On Flickr30K image-to-text retrieval, for example, simply substituting the standard image encoder with our ENN-trained version elevates the R@1 score from 87.9\% to 93.9\%, which is a remarkable 6.0\% absolute gain. Similarly, on COCO text-to-image retrieval, R@1 improves by \textbf{9.3\%} absolute. To be more specific, according to Figure~\ref{fig:it_visual}, the structured, decorrelated visual features learned by ENN, such as distinct representations for ``motorcycle," ``red helmet," and the number ``five," are fundamentally more general. They can be more cleanly and unambiguously mapped to their corresponding semantic concepts in the textual domain, reducing noise and interference in the shared embedding space. This confirms that ENN fosters a ``generalization flywheel," where enhanced unimodal representation quality directly translates to state-of-the-art performance on complex, cross-modal tasks that require a deep understanding of both vision and language.

\subsection{BP-Free Local Learning}
We next examined a fully BP-free training mode of our approach, ENN-$\ell$, to demonstrate its performance under local learning and the resulting parallelism benefits. In this setting, the network is divided into blocks and trained with our eigenvector loss block-by-block, without propagating gradients backward through the entire network. This eliminates the locking problems of standard BP, enabling concurrent updates. We evaluated ENN-$\ell$ on the same suite of image classification benchmarks described earlier and compare it to recent state-of-the-art local-learning methods.

\textbf{Classification Performance.} Remarkably, the results in Table~\ref{tab:img_class} reveal that ENN-$\ell$ not only outperforms all compared BP-free methods but also surpasses standard end-to-end BP on nearly every benchmark. On CIFAR-10, ENN-$\ell$ consistently shows lower error rates than BP. Compared with the strongest previous BP-free models, ENN-$\ell$ still outperformed with all backbones. A similar pattern is observed on CIFAR-100: ENN-$\ell$ reduces the BP baseline by around 4\% for all backbones while surpassing the second-best BP-free model, BWBPF, by over 2\% across the same depths. Of note, ENN-$\ell$ maintains a nearly constant error as depth increases—the gap between ResNet50 and ResNet152 is only 0.38\% on CIFAR-10, indicating that the eigenvector-based loss remains well-conditioned even in deep residual networks.

This trend holds across all datasets. On the Tiny ImageNet dataset, ENN-$\ell$ shows superiority to the traditional BP method and other BP-free models. For instance, ENN-$\ell$ achieves a 33.75\% error rate for ResNet-50, which is the lowest, outperforming both BP over \textbf{10\%} and the second-best BP-free baseline, BWBPF, for nearly 2\%. On the full ImageNet benchmark, ENN-$\ell$ delivers substantial accuracy gains over both BP and prior local-learning methods. For example, with a ResNet-101 backbone, ENN-$\ell$ reduces the top-1 error for nearly \textbf{20\%} of standard BP. Compared to the strongest published BP-free approach, BWBPF, ENN-$\ell$ also cuts error by 3\% on these ImageNet models. While BWBPF attains a slightly better result on the ViT-B/16 architecture, the margin is under 0.6 \%, showing that ENN-$\ell$ transfers effectively to transformer-based models without any architecture-specific tuning. These exhaustive results demonstrate that ENN's structured parameterization enables a local learning rule that is not only efficient but is consistently more effective than global error propagation.

\begin{table}[h]
\centering
\footnotesize
\begin{tabular}{@{}c|ccc@{}}\toprule
\textbf{Method} & \textbf{ResNet-101} & \textbf{ResNet-152} & \textbf{ViT-B/16} \\
\midrule
BP & 27.60  & 40.72 & 55.56 \\
DGL & 14.38  & 18.26 & -- \\
SEDONA$^{\dagger}$ & 13.73  & 20.16  & -- \\
BWBPF & 13.33  & 18.10  & 24.02  \\
ENN-$\ell$ & \textbf{13.10} & \textbf{18.06} & \textbf{23.17} \\
\bottomrule
\multicolumn{4}{l}{SEDONA$^{\dagger}$ results as reported in its paper.}
\end{tabular}
\caption{Training time ($\downarrow$) in hours on ImageNet, compared across different methods for ResNet-101, ResNet-152, and ViT-B/16. All models are split into 4 blocks and trained on 4 GPUs for 90 epochs. ENN achieves the shortest training time in every setting.}
\label{tab:time}
\end{table}

\textbf{Distributed Training Efficiency.} Beyond accuracy, a primary motivation for BP-free learning is to unlock model parallelism. We measure wall-clock training time on ImageNet by distributing models across 4 NVIDIA A100 GPUs. For a fair comparison, all methods were run for 90 epochs with a global batch size of 256. 

As depicted in Table~\ref{tab:time}, ENN-$\ell$ is unequivocally the fastest training method in every configuration. ENN-$\ell$ completes ImageNet training in 13.10 hours on ResNet-101 and in 18.06 hours on ResNet-152. On the other hand, traditional BP takes 27.6 hours on ResNet-101 and 40.72 hours on ResNet152. This corresponds to a greater than $2\times$ speedup in practice. Even with the ViT-B/16 transformer, ENN-$\ell$ finishes in 23.17 hours, substantially faster than BP. 

Table~\ref{tab:speedup} provides a deeper analysis of this efficiency. ENN-$\ell$ achieves the highest speedup ratios relative to BP in all cases: 2.11$\times$ for ResNet-101, 2.25$\times$ for ResNet-152, and 2.40$\times$ for ViT-B/16. Crucially, this state-of-the-art speedup is achieved with virtually no increase in model parameters. For instance, the ENN-$\ell$ ResNet-152 has 60.22M parameters, almost identical to the 60.19M of the BP baseline. This contrasts sharply with methods like SEDONA, which requires 86.00M parameters to achieve a lower speedup of 2.02$\times$. These results confirm that ENN-$\ell$ translates theoretical parallelism into substantial, practical gains in training efficiency without the cost of parameter bloat.

\begin{table}[h]
    \centering
    \footnotesize
    \begin{tabular}{@{}c|c|cc@{}}\toprule
    Model & Method & Speedup Ratios & Params. (M) \\
    \midrule
    \multirow{4}{*}{ResNet-101}
    & BP & 1 & 44.55 \\
    & DGL & 1.92 & 47.09 \\
    & SEDONA$^{\dagger}$ & 2.01 & 70.36 \\
    & BWBPF & 2.07 & 46.34 \\
    & ENN-$\ell$ & \textbf{2.11} & 44.61 \\
    \midrule
    \multirow{4}{*}{ResNet-152}
    & BP & 1 & 60.19 \\
    & DGL & {2.23} & 62.73 \\
    & SEDONA$^{\dagger}$ &  2.02 & 86.00 \\
    & BWBPF & 2.25 & 61.98 \\
    & ENN-$\ell$ & \textbf{2.25} & 60.22 \\
    \midrule
    \multirow{2}{*}{ViT-B-16}  & BP  & 1 & 86.86 \\
    & BWBPF & 2.31 & 88.57 \\
    & ENN-$\ell$ &  \textbf{2.40} & 86.97 \\
    \bottomrule
    \multicolumn{4}{l}{$^{\dagger}$Results are from SEDONA.}
    \end{tabular}
    \caption{Parallel training speedup ratios ($\uparrow$), and the number of parameters ($\downarrow$) compared among different methods on ResNet families and ViT, each distributed to 4 GPUs, when applied to \textbf{ImageNet}.}
    \label{tab:speedup}
\end{table}

\subsection{Embedding ENN-$\ell$ in 3D-CNN for Brain MRI Analysis}
As a final capstone, we demonstrated that ENN-$\ell$ is not a brittle solution tailored for 2D natural images but a robust, general-purpose framework that scales to entirely different data domains and dimensions. We apply ENN-$ell$ to the task of brain age prediction from 3D MRI scans, integrating it into a 3D-CNN~\cite{yin2023anatomically}.

\textbf{Experimental Setup.} The 3D CNN we used is configured with 4 sequential blocks, each trained locally with ENN-$\ell$. We train the model on a large brain MRI dataset comprising 4,681 subjects, and evaluate on an independent set of 1,170 subjects drawn from multiple cohorts (including UK Biobank, HCP-Aging, HCP-Young Adult, and ADNI). The network outputs a predicted age for each brain MRI, and performance is measured by the mean absolute error (MAE) between predicted age and true age (in years). We compare the ENN-trained 3D network to an equivalent 3D CNN trained with standard end-to-end backpropagation. 

\textbf{Results and Implications.} Even in this challenging high-dimensional setting, ENN-$\ell$ maintains strong performance while conferring efficiency benefits. The ENN-based 3D model achieves an MAE of 2.11 years on brain age prediction, which is notably lower than the 2.41 years MAE obtained by the backprop-trained 3D CNN. In other words, ENN-$\ell$ improves prediction accuracy by reducing error by about \textbf{12\%} in this task. While this is an initial exploration, the improved accuracy and efficiency on brain age prediction suggest that ENN-$\ell$’s advantages are not limited to conventional computer vision tasks, but also carry over to real-world domains that demand processing of high-dimensional data.

At the same time, training a 3D CNN with ENN-$\ell$ is significantly more efficient: thanks to block-wise parallelism and alleviating the backpropagation memory bottleneck, the 3D CNN with ENN-$\ell$ trains faster and uses less memory than its BP counterpart. This case study demonstrates that our approach effectively scales to complex biomedical imaging applications. By enabling efficient learning on 3D data, ENN-$\ell$ can make it practical to train large models on volumetric medical scans, potentially accelerating research in areas like neuroimaging. 

\section{Conclusion}
In this work, we confronted a foundational pathology of deep learning: the tendency of gradient-based optimization to produce disordered weight structures that degrade feature quality and hamper training in procedural bottlenecks. We introduced the Eigen Neural Network (ENN), an architectural paradigm that resolves this pathology not through post-hoc regularization, but by architectural axiom. Our comprehensive experiments demonstrate that this single architectural prior yields a cascade of benefits. When integrated into standard backpropagation, ENN's principled structure consistently produces more discriminative features, leading to state-of-the-art performance on large-scale image classification benchmarks. These superior unimodal representations lead to generalization, transferring seamlessly to set a new benchmark in the complex domain of cross-modal image-text retrieval. Most significantly, the eigenbasis representation unlocks a fully parallelizable, backpropagation-free local learning regime, ENN-$\ell$, that breaks the long-held compromise between training efficiency and model accuracy. ENN-$\ell$ not only resolves BP's locking constraints to achieve over a 2$\times$ training speedup but also consistently surpasses the accuracy of its end-to-end, globally trained counterparts.

By tackling the root cause of weight disorder, ENN offers a unified framework for designing models that are simultaneously more accurate, more efficient, and more generalizable. This work obviates the trade-off between the performance of backpropagation and the efficiency of its alternatives, opening a new path toward architecturally principled deep networks. Future work might explore if learned eigenbasis can serve as universal, transferable priors for few-shot learning, or if the extreme parallelism of ENN-$\ell$ can unlock on-device training regimes for models of a scale previously thought infeasible.
\bibliography{aaai2026}
\clearpage
\newpage
\appendix
\section{Code and Data Availability}
Our code to run the experiments can be found at https://github.com/Belis0811/Eigen-Neural-Network. MRI data are publicly available from ADNI, UKBB, CamCAN, and HCP. There are no relevant accession codes required to access these data, and the authors had no special access privileges that others would not have to the data obtained from any of these databases. 

\section{Training Algorithms}

\textbf{ENN-$\ell$ local training.} We outlined the layer-local learning procedure in \textbf{Algorithm~\ref{alg:enn}} that enablesENN-$\ell$ to train without global back-propagation. Each weight matrix is reconstructed on-the-fly from its orthonormal factors $Q^{(\ell)}$ and $P^{(\ell)}$ and a diagonal eigenvalue vector $\Lambda^{(\ell)}$.  During the forward sweep, the hidden representation $\mathbf{h}^{(\ell)}$ is \emph{detached} before it is consumed by the next layer, thereby preventing gradients from flowing across layers. A shallow head $h^{(\ell)}$ attached to every depth provides a layer-specific prediction $\hat{\mathbf y}^{(\ell)}$, yielding a cross-entropy term $L_{\mathrm{cls}}^{(\ell)}$ that drives local supervision.

To stabilise the factorisation, we impose an orthogonality penalty $L_{\mathrm{orth}}^{(\ell)}$ that keeps the columns of $Q^{(\ell)}$ and $P^{(\ell)}$ close to orthonormal.  The total loss for each layer is a simple weighted sum $L^{(\ell)} = L_{\mathrm{cls}}^{(\ell)} + \lambda\, L_{\mathrm{orth}}^{(\ell)}$. Because $L^{(\ell)}$ depends only on parameters local to layer $\ell$, the gradients $\nabla_{(Q,\Lambda ,P)}L^{(\ell)}$ can be evaluated and applied fully in parallel across all layers. This eliminates the \emph{backward-locking} and \emph{update-locking} bottlenecks of conventional back-propagation and allows asynchronous optimisation on distributed hardware.

\begin{algorithm}[tb]
\caption{Local Training of ENN-$\ell$} 
\label{alg:enn}
\begin{algorithmic}[1]
\STATE \textbf{Input:} Mini–batch $(\mathbf{x},\mathbf{y})$,
       layers $\ell=1,\dots,L$ with factors $\bigl(Q^{(\ell)},\Lambda^{(\ell)},P^{(\ell)}\bigr)$,
       local heads $\{h^{(\ell)}\}$, learning rate $\eta$, orthogonality weight $\lambda$.
\STATE \textbf{Forward / detaching:}
       $\mathbf{h}^{(0)}\!\gets\!\mathbf{x}$.
       \FOR{$\ell=1$ \textbf{to} $L$}
          \STATE $W^{(\ell)} \!\gets\! Q^{(\ell)}\operatorname{diag}\!\bigl(\Lambda^{(\ell)}\bigr) P^{(\ell)\top}$
          \STATE $\mathbf{z}^{(\ell)} \!\gets\! W^{(\ell)} \mathbf{h}^{(\ell-1)}$;\,
                 $\mathbf{h}^{(\ell)} \!\gets\! \phi\!\bigl(\mathbf{z}^{(\ell)}\bigr)$
          \STATE $\hat{\mathbf{y}}^{(\ell)} \!\gets\! h^{(\ell)}\!\bigl(\mathbf{h}^{(\ell)}\bigr)$ 
          \STATE detach $\mathbf{h}^{(\ell)}$ before feeding to layer $\ell\!+\!1$
       \ENDFOR
\STATE \textbf{Layer–local losses:}
       \FOR{$\ell=1$ \textbf{to} $L$ \textbf{in parallel}}
          \STATE $L_{\text{cls}}^{(\ell)}\!=\!\mathcal{L}_{\text{CE}}\!\bigl(\hat{\mathbf{y}}^{(\ell)},\mathbf{y}\bigr)$
          \STATE $L_{\text{orth}}^{(\ell)}\!=\!
                 \lVert Q^{(\ell)\!\top}Q^{(\ell)}\!-\!I\rVert_F^{2}\!+\!
                 \lVert P^{(\ell)\!\top}P^{(\ell)}\!-\!I\rVert_F^{2}$
          \STATE $L^{(\ell)} \!=\! L_{\text{cls}}^{(\ell)} + \lambda\,L_{\text{orth}}^{(\ell)}$
       \ENDFOR
\STATE \textbf{Local parameter updates:}
       \FOR{$\ell=1$ \textbf{to} $L$ \textbf{in parallel}}
          \STATE $\bigl(Q^{(\ell)},\Lambda^{(\ell)},P^{(\ell)}\bigr)
                 \;\leftarrow\;
                 \bigl(Q^{(\ell)},\Lambda^{(\ell)},P^{(\ell)}\bigr) - \eta\,\nabla_{(Q,\Lambda,P)}L^{(\ell)}$ 
       \ENDFOR
\STATE \textbf{Return} updated parameters and losses $\{L^{(\ell)}\}_{\ell=1}^{L}$.
\end{algorithmic}
\end{algorithm}

In practice, a small orthogonality weight $\lambda$ (e.g.\ $10^{-3}$) is sufficient to maintain stable training, and standard optimisers such as Adam or SGD can be applied independently to each factor set $(Q^{(\ell)},\Lambda^{(\ell)}, P^{(\ell)})$.  The factorised form also reduces memory usage: storing two tall orthonormal matrices and a diagonal vector requires $\mathcal{O}(d^2)$ parameters, while the diagonal eigenvalues add only $\mathcal{O}(d)$ overhead. Empirically, we observe no degradation in accuracy relative to full back-propagation, but a substantial reduction in wall-clock time thanks to parallelised layer updates.

\textbf{ENN BP training.} We also demonstrate our ENN integration of traditional backpropagation. \textbf{Algorithm~\ref{alg:enn_bp}} trains ENN layers with ordinary back-propagation. Unlike the layer-local variant (Algorithm~\ref{alg:enn}), activations are \emph{not} detached between layers; a single cross-entropy loss is computed at the network head, and the resulting gradient signal is propagated through the entire stack. Orthogonality regularisation is still applied to every factor pair $(Q^{(\ell)}, P ^{(\ell)})$ to preserve numerical stability and constrain the weight spectrum.

\begin{algorithm}[tb]
\caption{End-to-End Back-Propagation Training of ENN}
\label{alg:enn_bp}
\begin{algorithmic}[1]
  \STATE \textbf{Input:} Mini–batch $(\mathbf{x},\mathbf{y})$; depth $L$; factor parameters
         $\{Q^{(\ell)},\Lambda^{(\ell)},P^{(\ell)}\}_{\ell=1}^{L}$; global head
         $h^{(L)}$; learning rate $\eta$; orthogonality weight $\lambda$.
  \STATE \textbf{Forward pass:} $\mathbf{h}^{(0)} \leftarrow \mathbf{x}$.
  \FOR{$\ell = 1$ \textbf{to} $L$}
      \STATE $W^{(\ell)} \leftarrow Q^{(\ell)}
             \operatorname{diag}(\Lambda^{(\ell)}) P^{(\ell)\top}$
      \STATE $\mathbf{z}^{(\ell)} \leftarrow W^{(\ell)} \mathbf{h}^{(\ell-1)}$,\;
             $\mathbf{h}^{(\ell)} \leftarrow \phi\!\bigl(\mathbf{z}^{(\ell)}\bigr)$
  \ENDFOR
  \STATE \textbf{Task loss:} $\hat{\mathbf{y}} \leftarrow h^{(L)}\!\bigl(\mathbf{h}^{(L)}\bigr)$,\;
         $L_{\mathrm{task}} \leftarrow \mathcal{L}_{\mathrm{CE}}(\hat{\mathbf{y}},\mathbf{y})$.
  \STATE \textbf{Orthogonality loss:}
         $L_{\mathrm{orth}} \leftarrow \sum_{\ell=1}^{L}
         \bigl\|Q^{(\ell)\top}Q^{(\ell)}\!-\!I\bigr\|_{F}^{2} +
         \bigl\|P^{(\ell)\top}P^{(\ell)}\!-\!I\bigr\|_{F}^{2}$.
  \STATE \textbf{Total loss:} $L \leftarrow L_{\mathrm{task}} + \lambda\,L_{\mathrm{orth}}$.
  \STATE \textbf{Back-propagation:} Compute $\nabla_{(Q,\Lambda,P)}L$ for all layers via the chain rule.
  \STATE \textbf{Parameter update:}
         $\bigl(Q^{(\ell)},\Lambda^{(\ell)},P^{(\ell)}\bigr) \leftarrow
         \bigl(Q^{(\ell)},\Lambda^{(\ell)},P^{(\ell)}\bigr) -
         \eta \,\nabla_{(Q,\Lambda,P)}L\quad\forall\,\ell$.
  \STATE \textbf{Return:} Updated parameters and loss $L$.
\end{algorithmic}
\end{algorithm}

Because the factorised weight form is agnostic to the surrounding architecture, the same routine supports a wide range of backbones. In convolutional networks such as ResNet, standard \texttt{Conv2d} and \texttt{Linear} layers are replaced by \texttt{ENNConv2d} and \texttt{ENNLinear}; in Vision Transformers, \texttt{ENNLinear} can parameterise both the patch-projection layer and the query/key/value projections inside self-attention blocks.  The global loss encourages coordination across layers, which is beneficial for tasks that profit from deep feature interaction, while the factor decomposition retains the memory and FLOP reductions characteristic of ENN.

\section{More Experiments and Discussion}
\subsection{Training Hyperparameter}
\textbf{Image Classification.} For CIFAR experiments, we train for the equivalent of $\sim$500 epochs with batch size 128; the initial learning rate is 0.001. On the larger Tiny ImageNet dataset, models are trained for $\sim$800 epochs with batch size 256, using a cosine annealing schedule from 0.001 down to 1e-5. Weight decay of $1\times10^{-4}$ is applied for ResNet-50/101, and $2\times10^{-4}$ for ResNet-152 to mitigate overfitting. For ImageNet training, the model is trained for 1500 epochs with a batch size of 256. The learning rate starts at 0.001 (0.005 for ResNet-152) and decays by 0.1 at scheduled intervals to $1\times10^{-4}$.

\textbf{Image Retrieval.} For MS-COCO, we train for $\sim$140 epochs with a global batch size of 512 (8$\!\times\!$64 image–text pairs) under \texttt{AdamW}, linearly warming the learning rate to $3\times10^{-4}$ during the first 2{,}000 steps and then cosine-annealing it to $1\times10^{-6}$. Weight decay is set to $5\times10^{-2}$ for the ENN–ViT/B-16 image encoder and $1\times10^{-4}$ for the projection head. Because \textbf{Flickr30K} is smaller, we extend training to $\sim$200 epochs, halve the peak learning rate to $1\times10^{-4}$, and retain the same decay schedule to mitigate over-fitting. Throughout, the text tower remains frozen; ENN blocks are regularized with an orthonormality penalty of $2\times10^{-4}$, and gradients are clipped at 1.0 for stability, in line with heuristics from. Input images are resized to $224^{2}$ and augmented with RandAugment ($N{=}2$, $M{=}9$), mix-up 0.1, and label smoothing 0.1.

\subsection{Ablation Studies}
\textbf{Choosing the Orthogonality Weight $\lambda$.} As defined in the loss function, the coefficient $\lambda$ balances the cross-entropy term $L_{\mathrm{CLS}}$ and the orthogonality regularizer $L_{\mathrm{orth}}$ that encourages $Q^{(\ell)}$ and $P^{(\ell)}$ to remain orthonormal.
We swept $\lambda$ in the interval $[0,0.01]$, a range
commonly recommended for spectral penalties in CNNs and ViTs~\cite{yoshida2017spectral,bansal1810can},
and trained ENN-BP for 15~epochs on a 10\% ImageNet subset to obtain
a rapid proxy for full convergence.

\begin{figure}[t]
    \centering
    \includegraphics[width=.8\linewidth]{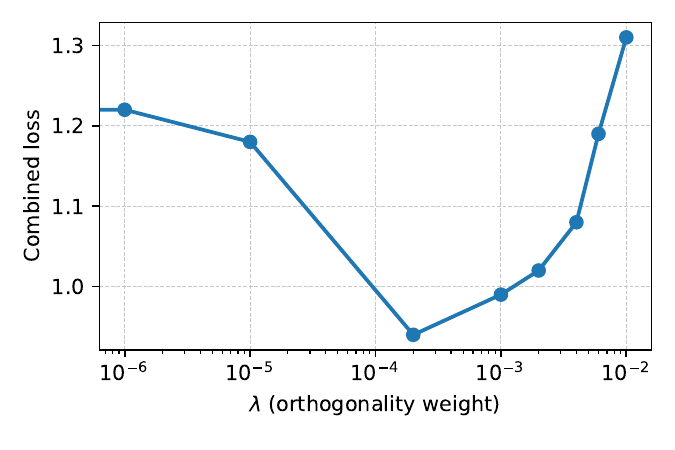}
    \caption{\textbf{Effect of the orthogonality weight $\lambda$.}
    Minimum validation loss is achieved at $\lambda=0.0002$.}
    \label{fig:lambda_orth}
\end{figure}

Figure~\ref{fig:lambda_orth} shows that the \emph{combined} loss
drops monotonically until $\lambda=0.0002$ and then rises again. If $\lambda$ is too small ($<\!10^{-5}$), orthogonality is under-regularized, leading to ill-conditioned weight factors and slower optimization.

Conversely, values above $0.004$ over-penalize $L_{\mathrm{orth}}$, dampening updates to $\Lambda^{(\ell)}$ and harming discriminative power. In subsequent experiments, we therefore fix $\lambda=0.0002$, which provides the best trade-off between fast convergence and robust eigen-basis conditioning.

Although the sweep spans only 15 epochs on a reduced dataset, the consistent U-shaped trend of the loss curve suggests that longer training or finer sampling is likely to refine the optimum only slightly. The chosen setting $\lambda=0.0002$ is therefore used throughout all large-scale experiments throughout the paper.

\textbf{Local Loss per layer.} To understand how each stage of ENN contributes to the final prediction, we attach auxiliary \emph{local heads} to the four main blocks of \textsc{ResNet-50}(conv\textsubscript{1}, conv\textsubscript{2\_x}, conv\textsubscript{3\_x}, and conv\textsubscript{4\_x}) and record their layer-wise cross-entropy losses during 20 epochs of training on CIFAR-10. This diagnostic follows the practice of layer-wise supervised or “early-exit’’ training~\cite{nokland2019training,lee2015deeply,szegedy2015going}, which shows that lower layers tend to learn generic features while upper layers focus on fine-grained semantics~\cite{chen2023layer,mostafa2018deep,bardes2021vicreg}.

\begin{figure}[t]
    \centering
    \includegraphics[width=.8\linewidth]{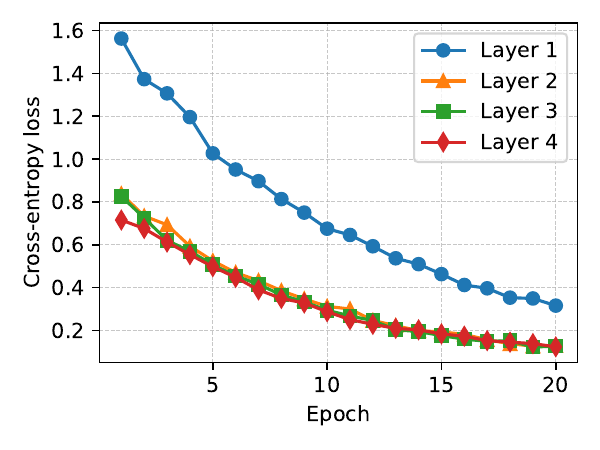}
    \caption{\textbf{Layer-wise training loss on CIFAR-10.}
    \textsc{ResNet-50} equipped with four auxiliary heads is
    trained for 20~epochs. Layer 1 (blue) starts highest;
    Layers 2 and 3 (orange/green) exhibit almost
    overlapping loss curves; Layer 4 (red) tracks Layer 3 with
    a marginal offset.}
    \label{fig:layer_loss}
\end{figure}

Figure~\ref{fig:layer_loss} reveals a consistent pattern: the loss at \textbf{Layer 1} starts highest and decays steadily yet remains above later layers, reflecting its direct exposure to the raw pixel distribution. \textbf{Layers 2} and \textbf{3} converge to nearly identical loss curves, indicating that their receptive fields capture similar mid-level abstractions~\cite{he2016deep,huang2017densely}. \textbf{Layer 4} maintains a loss slightly below (\emph{or occasionally above}) that of Layer 3, suggesting marginal additional discriminative capacity, but no over-specialization—an effect also reported in
orthogonally regularized CNNs~\cite{bansal1810can,he2024preventing}.

These observations confirm that the orthogonality-weighted local losses are well-balanced: the first stage learns rapidly without saturating deeper gradients, while deeper stages refine class-specific details.

The clear separation between the first block and the remaining ones aligns with prior findings that early convolutional layers encode generic edge/texture detectors, whereas intermediate groups share mid-level semantics before diverging again in the deepest layers ~\cite{zeiler2014visualizing,bau2017network}. Consequently, our local-loss formulation provides an intuitive diagnostic for monitoring convergence and adjusting the orthogonality weight if any stage lags behind.

\subsection{More comparison with the state-of-art BP-Free methods}
Table~\ref{tab:bp_free_comparison} contrasts \textbf{ENN-$\ell$} with leading BP-free training schemes on CIFAR-10 and CIFAR-100. Results for \emph{Direct Feedback Alignment(DFA)}~\cite{nokland2016direct}, \emph{Forward-Forward (FF)}~\cite{hinton2022forward}, \emph{NoProp}~\cite{li2025noprop}, and \emph{Mono-Forward(MF)}~\cite{gong2025mono} are taken from the original papers.

\begin{table}[h]
\centering
\small
\begin{tabular}{lcc}
\toprule
\textbf{Method} & \textbf{CIFAR-10 (\%↓)} & \textbf{CIFAR-100 (\%↓)}\\
\midrule
Forward-Forward  & 41.0 & -- \\
NoProp-DT     & 20.4 & 52.2 \\
DFA              & 10.6 & 38.6 \\
Mono-Forward & 38.2 & 45.2 \\
\midrule
\textbf{ENN-$\ell$ (ResNet-152)}             & \textbf{3.7} & \textbf{26.1}\\
\bottomrule
\end{tabular}
\caption{Test classification error (\% $\downarrow$) on CIFAR benchmarks. ENN-$\ell$ uses a ResNet-152 backbone; other numbers are the best test errors reported by the respective papers. “--” indicates that the corresponding paper did not report that result.}
\label{tab:bp_free_comparison}
\end{table}

Across both datasets, ENN-$\ell$ attains the lowest error among all BP-free methods. On CIFAR-10, it reduces error by \textbf{46\%} relative to the strongest published BP-free competitor, BDFA. On CIFAR-100, ENN-$\ell$ cuts error by more than \textbf{12\%} compared with BDFA and even more dramatically versus NoProp. These gains highlight the effectiveness of our eigenvector-based local learning rule in deep convolutional settings, demonstrating that BP-level accuracy can be matched, and in some regimes exceeded, \emph{without} global error back-propagation.

Qualitatively, the gap between ENN-$\ell$ and prior BP-free approaches widens as dataset complexity grows (from CIFAR-10 to CIFAR-100). We attribute this to ENN-$\ell$’s capacity to decouple learning of weight \emph{directions} and \emph{magnitudes}, allowing each layer to exploit richer local error signals than the scalar feedback employed by DFA, FF, and their variants. In sum, the evidence affirms ENN-$\ell$ as the current state-of-the-art for biologically inspired, fully parallel training on medium-scale vision benchmarks.
\end{document}